\documentclass[lettersize,journal]{IEEEtran}
\usepackage{amsmath,amsfonts}
\usepackage{algorithmic}
\usepackage{algorithm}
\usepackage{array}
\usepackage{textcomp}
\usepackage{stfloats}
\usepackage{url}
\usepackage{verbatim}
\usepackage{graphicx}
\usepackage{cite}
\usepackage{subcaption}
\usepackage{xcolor}
\usepackage{threeparttable}
\usepackage{booktabs}

\begin{document}

\title{Learning to Pick: A Visuomotor Policy for Clustered Strawberry Picking}

        % <-this % stops a space

% \markboth{IEEE ROBOTICS AND AUTOMATION LETTERS (RA-L). © IEEE}%

\author{Zhenghao Fei$^{1}$, Wenwu Lu$^{1}$, Linsheng Hou$^{1}$, Chen Peng$^{1}$$^{*}$  % <-this % stops a space
\thanks{1. ZJU-Hangzhou Global Scientific and Technological Innovation Center, Zhejiang University,
        Hangzhou, China
        {\tt\small \{zfei, chen.peng\}@zju.edu.cn}}%
\thanks{* corresponding author}
}

\maketitle

\begin{abstract}
Strawberries naturally grow in clusters, interwoven with leaves, stems, and other fruits, which frequently leads to occlusion. This inherent growth habit presents a significant challenge for robotic picking, as traditional percept-plan-control systems struggle to reach fruits amid the clutter. Effectively picking an occluded strawberry demands dexterous manipulation to carefully bypass or gently move the surrounding soft objects and precisely access the ideal picking point—located at the stem just above the calyx. To address this challenge, we introduce a strawberry-picking robotic system that learns from human demonstrations. Our system features a 4-DoF SCARA arm paired with a human teleoperation interface for efficient data collection and leverages an End Pose Assisted Action Chunking Transformer (ACT) to develop a fine-grained visuomotor picking policy. Experiments under various occlusion scenarios demonstrate that our modified approach significantly outperforms the direct implementation of ACT, underscoring its potential for practical application in occluded strawberry picking.
\end{abstract} 


\section{Introduction}
\IEEEPARstart{G}{lobal} demand for strawberries, a high-value crop, continues to rise. While China led production in 2023 with 3,336,690 tons, followed by the US with 1,055,963 tons \cite{Holmes31122024}, harvesting remains labor-intensive due to the fruit's fragility. This contrasts with mechanized harvesting of crops like corn and wheat. In both the US, where immigrant labor is crucial, and China, facing declining agricultural labor and rising costs, the need for robotic harvesting solutions is urgent for sustainable strawberry production.

% Given the difficulty of strawberry picking and no mature solution to date

Although numerous studies on robotic strawberry picking have been conducted  since the 1990s \cite{kondo1998strawberry}, no commercially viable system has yet been developed. These are main caused by the specialty of strawberries. Strawberries are particularly delicate, and their fragility means that they must be handled with great care—similar to the way human pickers operate. Additionally, strawberries grow in dense clusters, with leaves and stems often obscuring the target fruit, making manipulation among the cluster particularly challenging. A successful robotic picking system must not only identify the target fruit but also understand the entire scene, including the spatial relationships and physical properties of surrounding elements. This comprehensive perception is critical for developing a strategy that avoids obstacles, gently displaces soft and cluttered items, and accurately reaches the optimal picking point—just above the fruit’s calyx. The current state-of-the-art approach, which typically follows a percept-plan-control framework, struggles to address these complexities in scene understanding, depth estimation, trajectory planning, and control, limiting its effectiveness in real-world applications.
\begin{figure}[t]
\centering
    \includegraphics[width=0.5\textwidth]{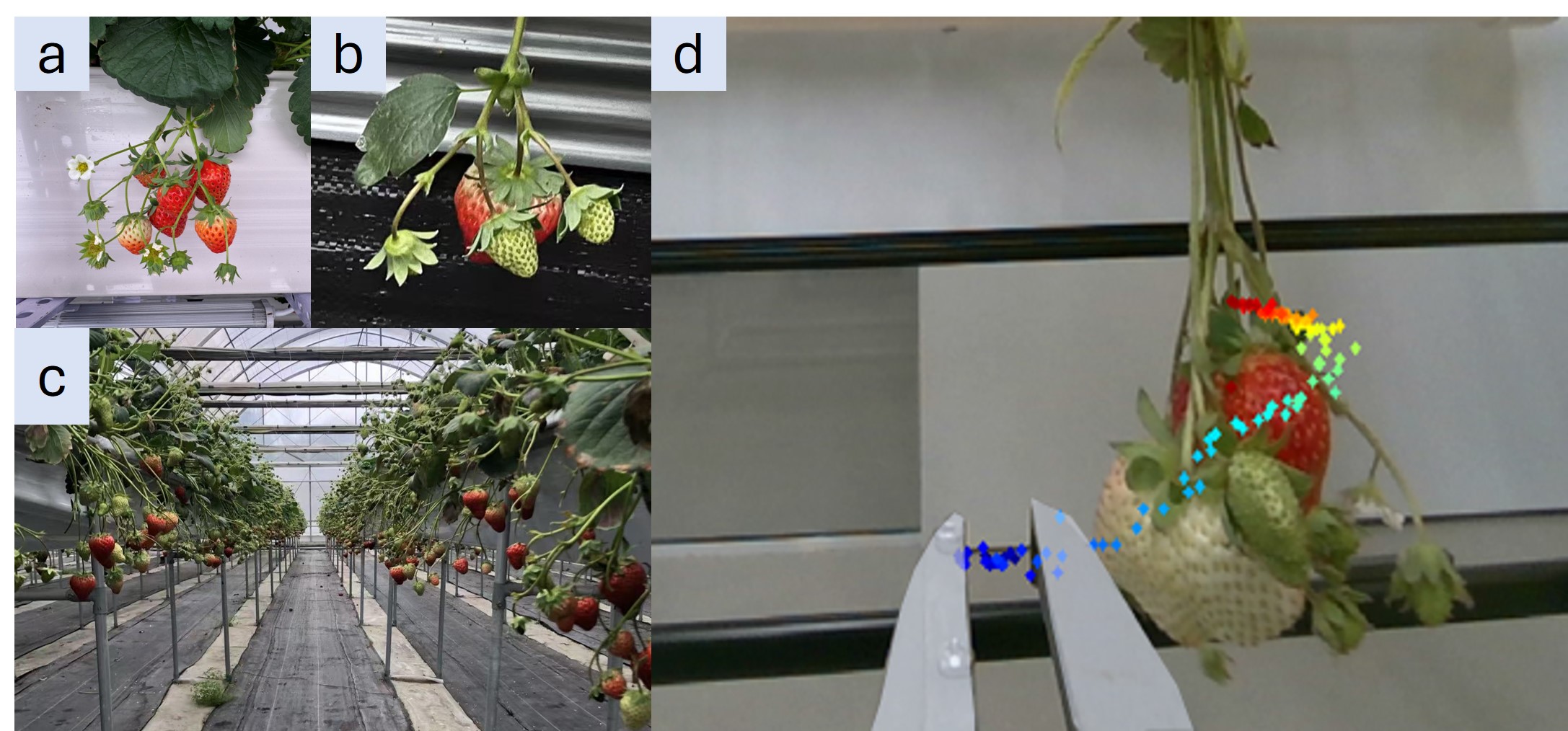}
    \caption{(a) \& (b) Real world challenge cases of clustered strawberries ; (c) Table-top strawberry cultivation; (d) Trajectory for dexterous picking for clustered strawberries;}
\label{fig:scenes}
\end{figure}

\begin{figure*} % Try [t!] instead of [h!]
  \centering
  \includegraphics[width=\textwidth]{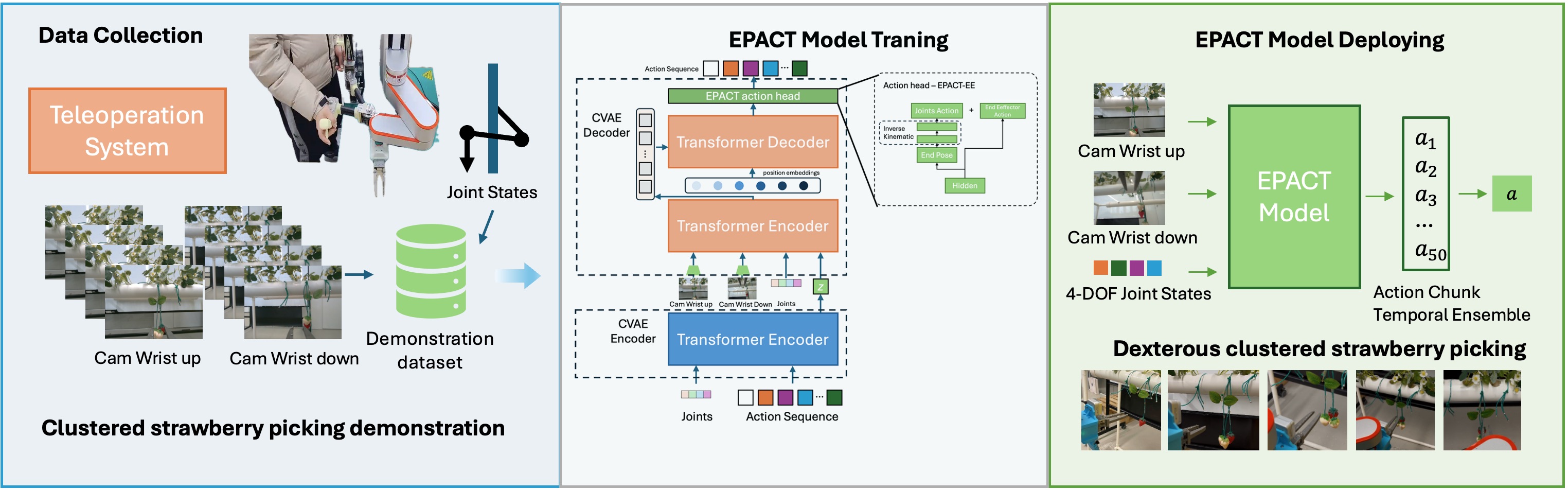} % Use \textwidth
  \caption{The overall framework of our work}
  \label{fig:headfigure}
\end{figure*}

Traditional robotic picking methods \cite{parsa2024modular,xiong2020autonomous}, which typically rely on precise modeling strategies, are not very fit to address the complexities of strawberry harvesting as mentioned above. The challenges posed by fragile fruit hidden among leaves, stems and unripe fruits, as well as the need for dexterous manipulation in cluttered environments, go beyond the capabilities of conventional techniques. These methods often fail to adapt to the dynamic nature of the task and the variability of the environment. In contrast, recent advancements in imitation learning techniques, such as the Action Chunking Transformer (ACT) \cite{zhao2023learning} and Diffusion Policy (DP) \cite{chi2023diffusion}, have shown promise in tasks that involve soft and non-rigid objects. For example, ACT has been used for cloth folding and shoe tying, while DP has been applied to dishwasher loading, both of which require delicate manipulation in cluttered environments. These methods, by learning from a few human demonstrations, might offer the flexibility and adaptability needed to handle the complexities of strawberry picking. Unlike traditional approaches, which struggle to model every aspect of the environment, imitation learning can focus on learning the core skills required for successful strawberry harvesting, providing a more promising solution.

% - Introduce what we want to do -
In our work, we propose a learning-based robotic system tailored to the specific challenges of strawberry picking. Strawberries typically grow in a vertical arrangement, which allows a 4-DoF robotic arm, such as a Selective Compliance Articulated Robot Arm (SCARA) \cite{tortuga}, to perform the necessary picking tasks effectively \cite{xiong2020obstacle}. Rather than relying on complex, high-degree-of-freedom manipulators that replicate the full range of human motion, our system utilizes a cost-effective 4-DoF design optimized for efficiency in commercial applications. A dedicated human teleoperation interface is used to collect demonstration data, and an End Pose Assisted Action Chunking Transformer method—built on the ACT algorithm—is developed to learn a dexterous visuomotor picking policy from these demonstrations.  The overall framework of our work is shown in Figure \ref{fig:headfigure}. Laboratory experiments under various occlusion conditions demonstrate that our learned policy outperforms the traditional percept-plan-control approach in picking occluded strawberries.

The remainder of this paper is organized as follows. In Section 2, related work of strawberry picking and imitation learning are reviewed. Section 3 describes our robotic strawberry picking system and the associated imitation learning approach. In Section 4, we detail the experimental setup and compare our method with the traditional percept-plan-pick pipeline. Finally, Section 5 concludes the paper and outlines directions for future research.

The contribution of this paper:

\begin{itemize}
 \item We developed a remotely operated data collection system for strawberry picking that integrates a specialized 4-DoF SCARA robotic arm with a human teleoperation interface. 
 \item We demonstrated the feasibility of a learning-based visuomotor policy for clustered strawberry picking using a 4-DoF Selective Compliance Articulated Robot Arm (SCARA), providing an effective alternative to common 6-DoF arms.
\item We demonstrate, to the best of our knowledge, the first successful learning and deployment of a dexterous visuomotor policy for picking real, clustered strawberries through imitation.
 \item We enhance the Action Chunking Transformer method by incorporating an end pose loss and a neural network based inverse-kinematic module, resulting in a success rate improvement for the picking tasks. 
 \item We conduct a comprehensive study to evaluate the impact of various configured occlusion scenarios on the performance of the proposed algorithm.

\end{itemize}

\section{Related Work}
% [Your content]
\subsection{Robotic Strawberry Picking}

Strawberry picking traditionally relies on a modular robotic manipulation pipeline \cite{ren2024mobile,xiong2020autonomous,parsa2024modular} that includes perception \cite{ge2019fruit}, planning, picking \cite{xiong2018design}, and placing. The perception system first detects the fruits and localizes their positions relative to the robot arm \cite{kim20232d}. Next, a planning system determines the picking order and generates trajectories to reach the target positions \cite{tafuro2022strawberry}. Finally, the end-effector executes the detaching operation and places the picked fruit into a collecting container \cite{xiong2020autonomous,han2012strawberry}. This classic pipeline fits the most cases of strawberry picking. However, there exist some challenging scenarios, particularly when fruits are clustered, which has led many researchers to introduce advanced techniques to address these issues \cite{xiong2020obstacle}.

Active perception techniques have emerged as effective solutions for overcoming occlusion challenges in robotic fruit picking \cite{rajendran2024towards}. These methods dynamically adjust the robot's viewpoint or sensor positioning to improve the visibility of target objects. For instance, Magistri et al. \cite{magistri2024improving} propose an active perception approach that integrates 3D shape completion with a 6-DoF pose estimation module. Using high-resolution point cloud data and a learned shape prior through DeepSDF, their system reconstructs the 3D geometry of occluded strawberries and tomatoes, enabling precise grasp point estimation and efficient end-effector placement. This approach significantly enhances picking performance in complex, real-world environments. Yi et al. \cite{yi2024view} propose a view planning approach for grape harvesting robots to overcome occlusion challenges, where fruit stems are hidden by leaves and poor observation angles. Their method uses an active vision strategy to alter the robot’s viewpoint, improving occlusion handling through a Spatial Coverage Rate Metric (SC) and incorporating motion cost. Experimental results on a real robot show that their approach achieves a higher picking success rate with reduced computation time compared to other advanced planners. However, the active-perception-based method requires extra computing power to process multiple views and adjust the robot's viewpoint, potentially increasing processing times. This trade-off between improved accuracy and longer computation time is a key consideration for real-time, large-scale agricultural applications.

Alternatively, an active obstacle separation strategy offers another method to tackle occlusion challenges by focusing on selectively picking a target fruit surrounded by obstacles. Xiong et al. \cite{xiong2020obstacle} utilizes 3D visual perception to guide push and drag motions, effectively separating obstacles from the target. The zig-zag push breaks static contact forces, while the drag operation relocates the target fruit to a less obstructed area for easier detachment. This method has demonstrated adaptability to complex scenarios and improved picking success rates in cluttered environments. Mghames et al. \cite{mghames2020interactive} present the Interactive Probabilistic Movement Primitives (I-ProMP) algorithm for path planning in fruit picking, addressing the challenge of pushing occluding unripe fruits to access ripe ones. While the approach offers computational efficiency and shows successful demonstrations in simulation, it has several limitations for real-world application. The method was validated only in a simulated environment, modeling strawberry stems as simple 3-axis hinges, which oversimplifies the complex, deformable properties of real plants. Furthermore, the I-ProMP algorithm operates as a planner that assumes perfect perception—presuming the 3D locations of all fruits are already known and that stem orientation is a given input—rather than addressing the more complex end-to-end visuomotor problem and the work does not demonstrate performance on a physical robot.

\subsection{Imitation Learning}

Learning-based methods, particularly those derived from imitation learning (IL), have shown great promise in enabling robots to learn complex tasks from demonstrations. However, traditional IL approaches often struggle with limited expressiveness when capturing the multi-modal representations needed for intricate manipulations, such as those required in clustered strawberry picking tasks. These methods typically learn a single deterministic mapping, which can be insufficient for addressing the variability and uncertainty inherent in real-world environments. In \cite{tafuro2022dpmp}, the authors present a novel probabilistic extension of deep movement primitives (DMPs) to address these limitations. By mapping visual information into a distribution of effective robot trajectories, this approach captures the multi-modal nature of complex manipulations.

Recently, imitation learning methods with advanced architectures have shown great potential for complex robotic manipulation. Action Chunking with Transformers (ACT) \cite{zhao2023learning,aldaco2024aloha,zhao2024aloha} improves upon traditional methods like deep probabilistic motion planning (dPMP) by addressing key limitations. While dPMP maps visual data to probabilistic trajectory distributions using static domain-specific latent spaces, it lacks mechanisms to handle temporal dependencies. ACT overcomes this with a transformer-based architecture and action chunking, allowing it to predict sequences of actions, reducing task horizon and mitigating compounding errors. Additionally, its use of temporal ensembling ensures smoother and more precise trajectories. By integrating spatial and temporal dependencies with closed-loop visual feedback, ACT provides dynamic adaptability to complex, real-world environments, outperforming static distribution-based approaches like dPMP.

Diffusion-based policies (DP), as demonstrated in \cite{chi2023diffusion}, bring additional versatility to imitation learning in agricultural contexts. The DP framework excels in handling multi-modal action distributions and scaling to high-dimensional action spaces, making it well-suited for challenging tasks such as outdoor pepper harvesting \cite{kim2024autonomous}. Unlike ACT, DP is easier to set up and collect data for, as shown in the Universal Manipulation Interface (UMI) framework \cite{chi2024universal}, which simplifies data collection through handheld devices equipped with fiducial markers. This enables large-scale data gathering in unstructured environments. In \cite{kim2025autonomous}, Kim et al. leverage the UMI-based method for autonomous pepper harvesting, demonstrating the feasibility of learning-based approaches for scalable robotic harvesting in dynamic field conditions. However, while DP is more accessible and generalizable, we chose the Action Chunking Transformer (ACT) for strawberry picking due to its compatibility with our 4-DOF robotic arm. ACT processes 4-DoF end-effector movements directly, aligning seamlessly with our system's capabilities. In contrast, Diffusion Policy (DP) requires recording 6-DoF end-effector motions. Projecting these 6-DoF movements to 4-DoF can introduce inaccuracies, potentially compromising the precision needed for delicate tasks like strawberry picking. Therefore, ACT's direct application to 4DoF movements makes it a more suitable choice for our specific robotic setup.

\section{Material and methods}
% [Your content]
\subsection{System Hardware}

\subsubsection{Picking Robot}

The robot arm employed in our system is a 4-DoF SCARA unit, selected for its suitability in horticultural environments—especially in the context of tabletop strawberry picking in vertical farming. Since strawberries in these settings are cultivated on horizontal platforms with limited vertical space, the SCARA arm’s planar motion and restricted degrees of freedom are ideally matched to the picking task. Its simplified geometry and compliance enable it to effectively navigate tight clusters of leaves and stems with minimal complexity. Moreover, SCARA arms are known for their rapid cycle times and precise, repeatable motions, which facilitate faster picking operations and improve overall picking efficiency.

\subsubsection{Human Teleoperation system}

To collect data from human demonstrations, we designed a remote tele-operation system. A human operator uses a twin-arm interface to directly control the SCARA arm. This system maps the rotational joint angles of the twin-arm interface, including the grippers, to the SCARA arm's corresponding joints. Joint angle encoders on the SCARA arm enable closed-loop control of these mapped positions. Two cameras, mounted on the end-effector (one top, one bottom), capture synchronized images and joint angle data during demonstrations. This synchronized data allows the robot to learn expert picking behaviors and provides high-quality training data for our end-to-end visuomotor policy.

\begin{figure}[t]
\centering
    \includegraphics[width=0.45\textwidth]{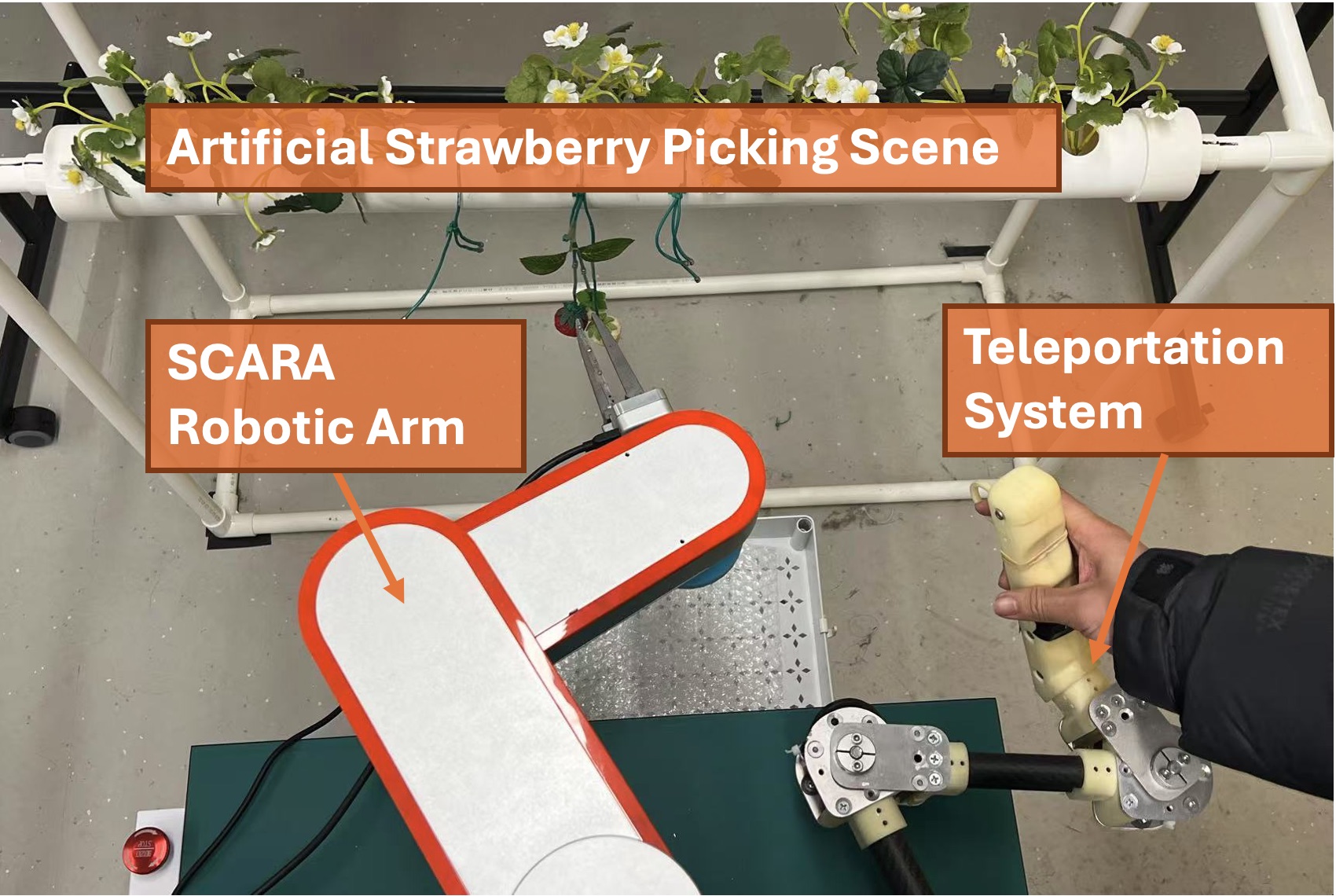}
    \caption{Illustration of a remotely operating system of SCARA for data collection.}
\label{fig:remotely_operating}
\end{figure}

\subsection{Methodology}
\subsubsection{Problem Formulation}

% As seen in related works \cite{xx}, recent breakthroughs in imitation learning have demonstrated significant potential in solving the manipulation of deformable objects, such as cloth folding and shoe lacing. Similarly, we believe that such learning-based visuomotor policies can benefit clustered strawberry picking, given that successfully picking an occluded strawberry in a tangled cluster involves the dexterous manipulation of deformable (biological) objects.
Recent advances in imitation learning have shown promise in manipulating deformable objects—such as cloth folding and shoe lacing—which require fine-grained, closed-loop control. Inspired by these breakthroughs, we address the challenging task of picking occluded, clustered strawberries using a learning-based visuomotor policy.

% In this work, we formulate the problem of picking clustered and tangled strawberries as a future action prediction problem based on current observations. A policy network πθ\pi_\theta predicts a sequence (or chunk) of future actions ˆat:t+k\hat{a}_{t:t+k} based on the current observations oto_t. The SCARA robot's actions are represented as the combination of 4-DOF arm joints position and a 1-DOF gripper position, denoted as a∈R5a \in \mathbb{R}^5. The observation oto_t includes single or multiple images from the wrist-mounted cameras combined with the robot's proprioceptive state. Summarizing, ot=[I1t,…,Int,qt]o_t = \left[I_t^1, \ldots, I_t^n, q_t\right] where IntI_t^n represents the nthn^{th} image at time tt and qtq_t is an array of each joint angle of the robot arm.

% The policy is trained through imitation learning to mimic human picking demonstrations collected using the tele-operation device introduced in this work. The optimization objective is to minimize the difference between the predicted action sequence and the demonstrated action sequence given the same observation.

We formulate the clustered strawberry picking task as a future action prediction problem based on current observations. At each time step $t$, the policy network $\pi_{\theta}$ observes the current state $o_t$ and generates a sequence (or chunk) of future actions $\hat{a}_{t:t+k}$. The robot’s action space consists of the 4-DOF joint positions of the SCARA arm and a 1-DOF gripper position, denoted as $a \in \mathbb{R}^5$. The observation $o_t$ comprises images from the monocular RGB camera(s) along with the robot’s proprioceptive state, summarized as:
$o_t = \left[I_t^1, \ldots, I_t^n, q_t\right]$, where $I_t^n$ denotes the $n^{th}$ image(s) at time $t$ and $q_t$ represents the vector of joint angles.

The policy is trained via imitation learning from human demonstrations collected with our remote tele-operation system. The training objective is to minimize the discrepancy between the generated action sequence from $\pi_{\theta}$ and the human demonstrated action sequence given the same observation. This formulation enables the robot to acquire the dexterous visuomotor skills necessary for accurately picking cluttered strawberries.

\subsubsection{End Pose Assisted Action Chunking with Transformers} 
% Specifically, we designed the EPACT (End Pose Assisted Action Chunking with Transformers) algorithm, which builds on the novel imitation learning method known as ACT (Action Chunking with Transformers), proposed by \cite{zhao2023learning}. The EPACT algorithm includes an additional focus on the accuracy of the robot arm’s end pose. The original ACT algorithm predicts robot actions in its joint space by minimizing the difference in each joint's angle relative to the demonstrated joint angles. Predicting actions directly in the joint space facilitates whole-body control and benefits highly dynamic actions [], however, even slight differences in joint angles can lead to significant discrepancies in the end pose, particularly for joints that are far from the end effector. Unlike cloth folding, picking clustered strawberries is exceptionally sensitive to the accuracy of the end pose trajectory. To enhance the focus of the imitation learning policy on the end pose trajectory, we proposed EPACT based on the original ACT without altering the policy training scheme.

Specifically, we introduce the EPACT (End Pose Assisted Action Chunking with Transformers) algorithm, which extends the imitation learning approach ACT (Action Chunking with Transformers) proposed by \cite{zhao2023learning}. The original ACT algorithm predicts robot actions in joint space by minimizing the difference between the generated and demonstrated joint angles. Although this approach facilitates whole-body control and benefits dynamic actions\cite{fu2023deep}, slight deviations in joint angles—especially in joints far from the end effector—can lead to significant errors in the end pose. Unlike cloth folding, picking clustered strawberries is exceptionally sensitive to the accuracy of the end pose trajectory. Our method enhances the policy’s emphasis on achieving the correct end pose without altering the underlying training scheme.

The main idea behind EPACT is that instead of using the policy network directly predicting each joint’s angle, it first predicts the end pose of the robot. Then, this end pose is decoded into individual joint angles through an "inverse kinematics" network.
The intermediate output, which consists of the end poses, is compared to the ground truth to formulate an end pose loss, denoted as $\mathcal{L}_{ep}$. Minimizing this loss during training encourages the policy to focus more on the accuracy of the end pose. Additionally, these intermediate end pose outputs can be used for visualizing predicted trajectories, providing a clearer understanding of the end-to-end policy’s intentions. 
To test this, two variants of the EPACT architecture were designed, differing specifically in the structure of their action prediction heads: EPACT-L (Latent) and EPACT-EE (End-Effector). The primary distinction lies in whether the end-effector (gripper) action is predicted jointly with the arm kinematics or in a decoupled manner. 

\begin{figure}[t]
\centering
    \includegraphics[width=0.48\textwidth]{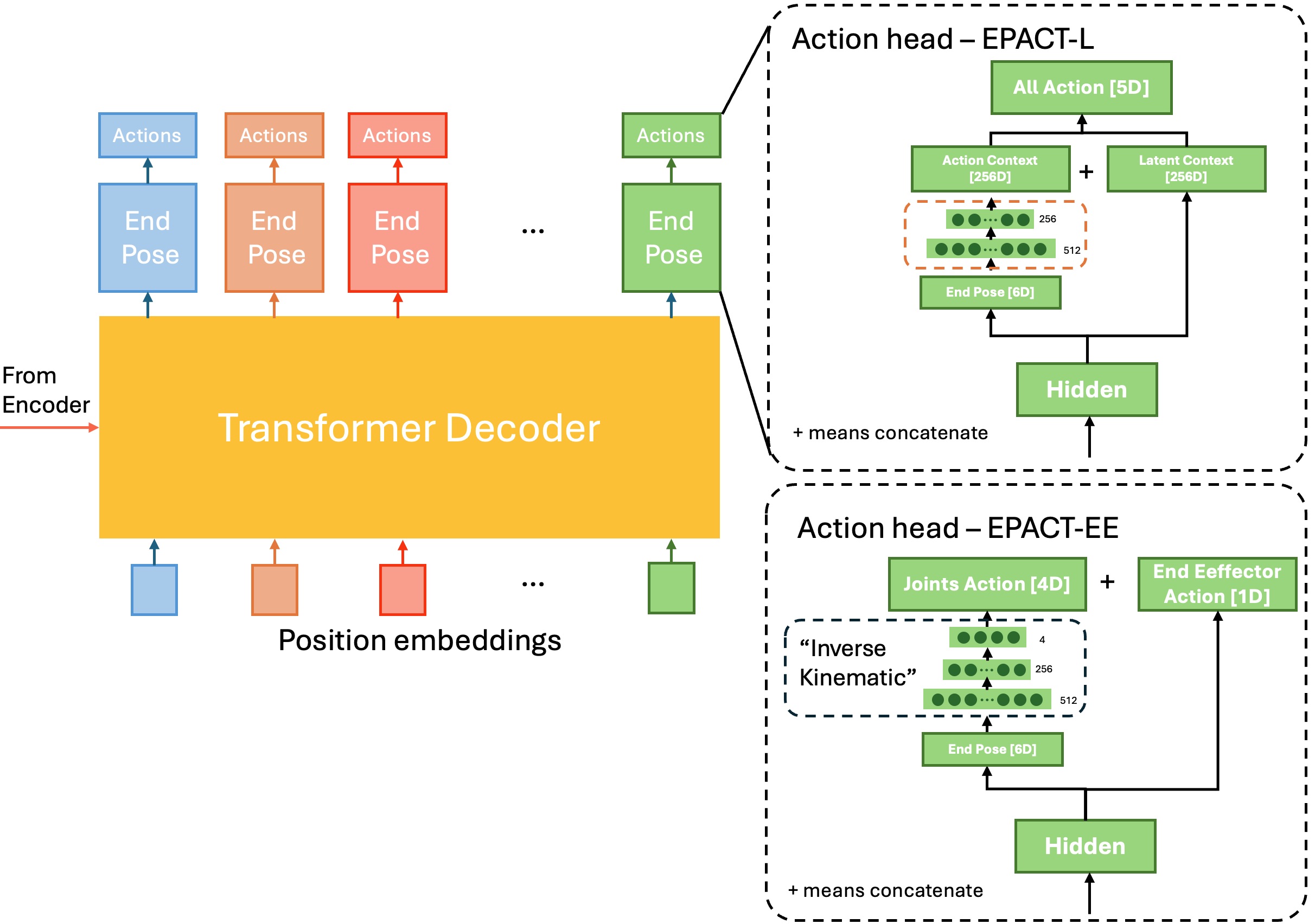}
    \caption{EPACT-L and EPACT-EE Action Prediction Decoder}
\label{fig:arch1}
\end{figure}

1. EPACT-L (Latent): As illustrated in Figure \ref{fig:arch1} EPACT-L, this variant uses a unified action head. It attempts to predict the entire 5D action vector (4 arm joints and 1 gripper state) from the predicted 6D end-effector pose. Because the gripper's open/close state cannot be directly inferred from an end-effector pose, this model implicitly relies on auxiliary information from the transformer's hidden state (latent context) to resolve the ambiguity and generate the complete action vector within a single neural network module. 

2. EPACT-EE (End-Effector): This variant, shown in Figure \ref{fig:arch1} EPACT-EE, was designed to address the logical challenge of the unified model. It features a decoupled action head with two parallel branches. Branch 1 (Arm Kinematics): The predicted 6D end-effector pose is fed into an neural
network based "Inverse Kinematic" module that is responsible only for decoding it into the 4D joint actions for the SCARA arm. Branch 2 (Gripper Action): In parallel, the gripper's 1D action is predicted directly from the shared hidden state of the transformer decoder, completely bypassing the end-pose calculation. The outputs from these two branches are then concatenated to form the final 5D action vector. This decoupled design hypothesizes that separating the arm's spatial goal from the gripper's binary action could lead to a more robust and learnable policy, as each branch focuses on a more clearly defined sub-tasks.

% Two variants of EPACT were designed. The first variant is called EPACT-L(Latent), Figure \ref{fig:arch1} illustrates the architecture of the EPACT-L’s action prediction decoder and action head. Figure \ref{fig:arch2} displays the architecture of the second variant's action head, which is denoted as EPACT-EE(End Effector). The primary distinction between the two variants lies in whether the action of the end effector is predicted independently from the joint actions. The rationale behind the design of EPACT-EE is that the action of the end effector cannot be derived from the end pose using inverse kinematics alone, as the gripper operation cannot be conducted from the end pose. In this decoupled setting, the "inverse kinematics" module learns the mapping from the end pose to the joint actions in an explicit way.

Besides the action head of the policy network's transformer decoder, the rest of the architecture remains the same as ACT. Readers can refer to \cite{zhao2023learning} for details.
%The complete architecture includes a Conditional Variational Autoencoder (CVAE) encoder and a decoder. During training, the CVAE encoder extracts a style variable from action sequences and joint observations; The CVAE decoder includes a transformer encoder, which combines one or two camera image observations with the joint observations and the style variable. Additionally, a transformer decoder is utilized to predict a sequence of actions.%

% \begin{figure}[t]
% \centering
%     \includegraphics[width=0.4\textwidth]{figures/arch2.jpeg}
%     \caption{EPACT-EE Action head}
% \label{fig:arch2}
% \end{figure}

The training objective for our policy includes minimizing three types of losses: the action reconstruction loss $\mathcal{L}_{rec\_action}$, the regulation loss $\mathcal{L}_{reg}$ that regularizes the CVAE encoder to a Gaussian prior, and the end pose reconstruction loss $\mathcal{L}_{rec\_end\_pose}$. The $\mathcal{L}_{rec\_action}$ and $\mathcal{L}_{reg}$ are kept the same with the original ACT method. The formulation of $\mathcal{L}_{rec\_end\_pose}$ is expressed as below.
\begin{equation}
    \mathcal{L}_{rec\_end\_pose} = L_1\left(\hat{e}_{t:t+k}, e_{t:t+k}\right)
\end{equation}
The formulation of the loss function during training is as follows:
\begin{equation}
    \mathcal{L} = \mathcal{L}_{rec\_action} + \beta\mathcal{L}_{reg} + \gamma\mathcal{L}_{rec\_end\_pose}
\end{equation}
where $\beta$ is the KL weight and $\gamma$ is the EP (end pose) weight. Here, $e_{t:t+k}$ represents a sequence of the robot arm's end poses, specified in terms of $\{x, y, z, \text{roll}, \text{pitch}, \text{yaw}\}$. These end poses are calculated from the joint states using the robot arm's forward kinematics. $\hat{e}_{t:t+k}$ denotes the predicted sequence of the robot arm's end poses.

\subsubsection{Data Collection}
We constructed an artificial strawberry picking scene, depicted in Figure \ref{fig:remotely_operating}, to mimic a tabletop strawberry farm environment with comparable visual complexity. Vertical cultivation pipes supporting artificial strawberry plants were used. Strawberry stems were simulated with flexible wires, and a magnetic system—small magnets on both the stem's end and the strawberry top—allowed for easy attachment and detachment. Red and white rubber strawberries were used to represent ripe and unripe fruit respectively, replicating the appearance and weight of real strawberries. This artificial scene allows for the creation of diverse strawberry clutter scenarios in a controlled lab setting, facilitating data collection and reproducible experiments.

For data collection, a human operator use the tele-operation arm to control the SCARA robot arm in a clustered strawberry picking task. Each demonstration commenced with the robot positioned facing a ripe(red) strawberry, but initial positions were intentionally randomized to ensure variety.. The operator then directed the robot to maneuver the gripper, either deviating around or gently pushing aside any obstructions, before securely grasping the strawberry stem and separating the fruit from the stem above the berry. This sequence of actions mirrors actual robotic strawberry harvesting, where robots are designed to cut and retain the stem above the fruit. Demonstration duration was approximately 10 seconds per instance

We designed five clustered strawberry picking scenarios to represent common cluster configurations found in the field, where each cluster consists of two or three strawberries. These scenarios mimic the difficulty of picking a target strawberry from such a cluster in actual field. We also included a non-clustered scenario for baseline verification. These scenarios, detailed in Figure \ref{fig:cluster_states}, were designed to mimic the difficulty of picking a target strawberry from within a cluster. Except for State 0, all scenarios introduce occlusions that obstruct a direct pick. While our artificial scene is more controlled than a real field, the inherent variability of stem deformation and leaf/flower placement prevents perfect replication of each state. Data was collected across these five scenarios, resulting in a total of 511 demonstrations.

\begin{figure}[t]
\centering
    \includegraphics[width=0.45\textwidth]{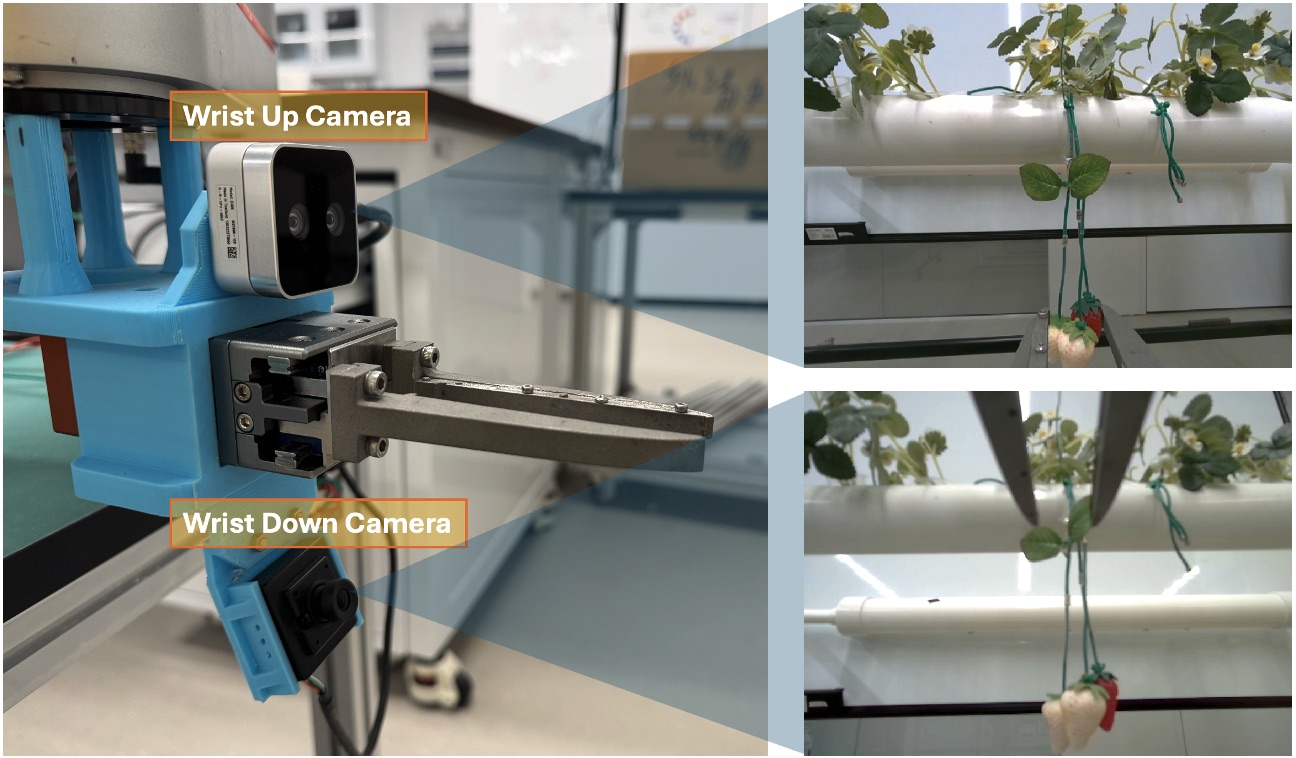}
    \caption{Camera setup; Images on the right represent the camera views}
\label{fig:wrist_cameras}
\end{figure}

Two wrist-mounted cameras are installed on the robot's end effector, one on top and one on the bottom, as shown in Figure \ref{fig:wrist_cameras}. The upper wrist camera is an Intel RealSense D405, and only its monocular RGB stream is used. The lower wrist camera is a low-cost USB web camera. During data collection, images from both cameras, denoted as $\left[I_t^{wrist-up}, I_t^{wrist-down}\right]$, are captured to provide comprehensive visual information from different views. This multi-view setup can help handling occlusions in the clustered strawberry picking task. The joint states encompass only the 4-DOF SCARA arm's joints, represented as $q \in \mathbb{R}^4$. Notably, the position of the gripper is excluded from these states. The target action data, denoted as $a \in \mathbb{R}^5$, includes command inputs for all joints obtained from the tele-operation device, with an additional value representing the gripper's opening state.

\subsubsection{Implementations}
Action chunking and temporal ensemble methods are also applied to enhance the stability of the robot arm's actions and to reduce motion jerkiness. To investigate the impact of camera number and viewpoint on the performance of occluded strawberry picking, we varied the cameras used in EPACT policy training and inference across three settings: \{Wrist Up\}, \{Wrist Down\}, and \{Wrist Up, Wrist Down\}. The image observations $\left[I_t^{wrist-up}\right]$, $\left[I_t^{wrist-down}\right]$, and $\left[I_t^{wrist-up}, I_t^{wrist-down}\right]$ were adjusted accordingly. The networks were trained using 511 demonstrations, among which 6 randomly selected demonstrations were fixed and used as validation data.

\section{Experiments}
\subsubsection{Evaluating Policies}
The experimental evaluation employed three network architectures: two EPACT variations and the ACT algorithm as baseline. Visual input configurations included utilizing all cameras mounted on the end-effector, as well as two separate monocular camera setups.

Consistent settings were applied across all policies for training and inference. Input camera images were resized to 480x640, and features were extracted using ResNet18 \cite{he2016deep}. The joint state was represented by a 4D vector of SCARA arm joint positions, and actions were 5D vectors (4 arm joints, 1 gripper). Policy predictions were generated at 30Hz, matching the camera frame rate, with an action chunking size of 100 and temporal ensembling enabled. Training utilized an NVIDIA RTX 4090 GPU, while inference was performed on an NVIDIA RTX 4050 Laptop GPU. The details of all policies are presented in Table \ref{tab:policy_comparison}.

\subsubsection{Experiments and results}
\begin{figure}
\centering
    \includegraphics[width=0.48\textwidth]{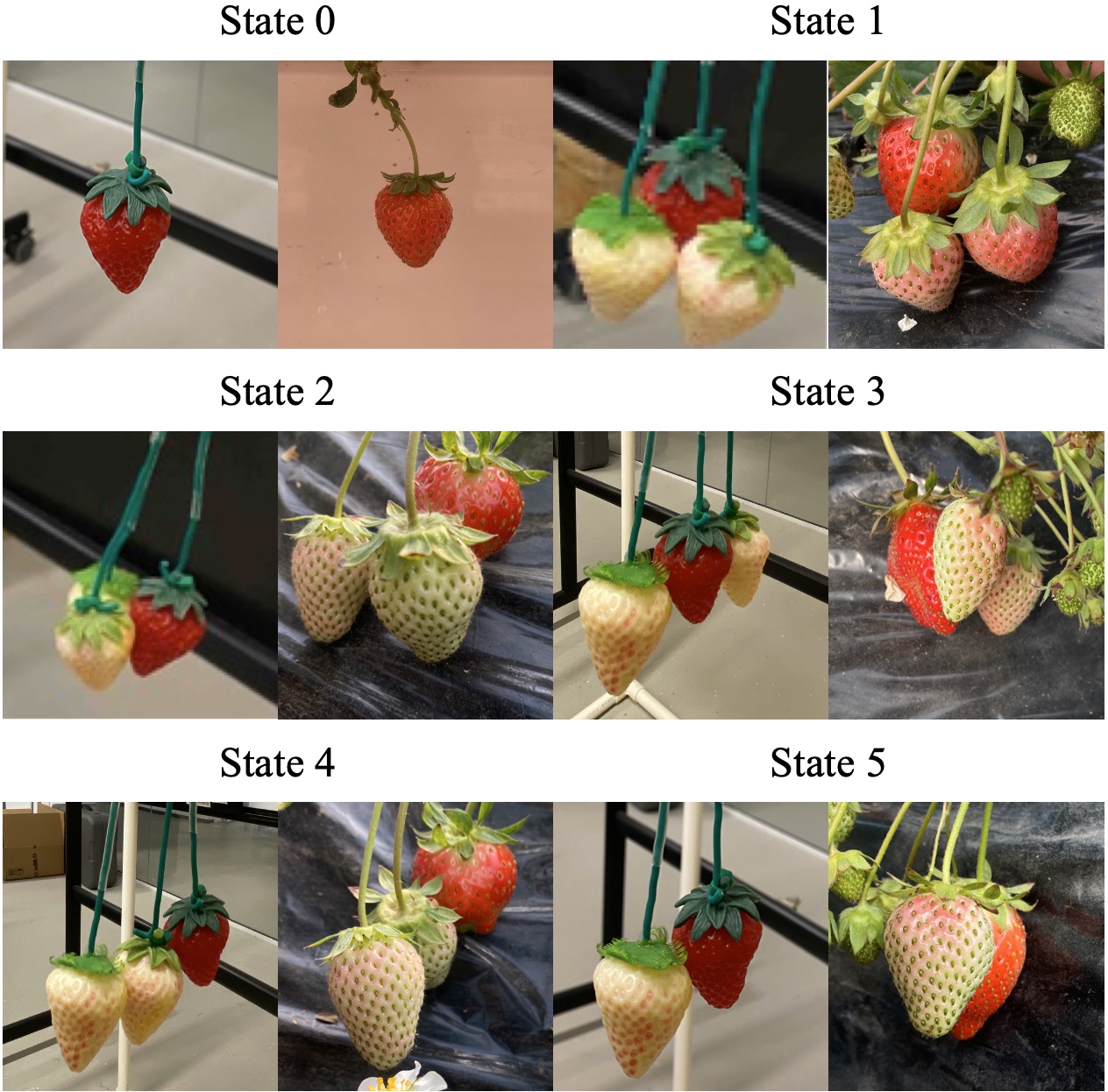}
    \caption{Four representative three‑strawberry cluster scenarios (States 1-4), a two-strawberries cluster (State 5) and a non-occluded strawberry baseline scenario (State 0). In each state, the left image shows the artificial scene used in our experiments, and the right image shows real strawberries in the field.}
\label{fig:cluster_states}
\end{figure}

We evaluated all policies across six strawberry cluster scenarios, detailed in Figure \ref{fig:cluster_states}. To account for scene variability, we performed 10 picking trials for each policy and scenario. Table \ref{tab:state_comparison} presents the successful picking rates achieved by each policy with full camera observation across all scenarios. The results clearly show that both EPACT variants outperformed the original ACT policy in all clustered states. Specifically, the end-pose assist module improved average success rates by 36.7\% for EPACT-L and 41.7\% for EPACT-EE. Given that all policies were trained on the same dataset and with identical configurations, these findings demonstrate the enhanced manipulation capabilities of the EPACT policy for clustered strawberry picking.

% \begin{table}[h]
% \centering
% \begin{tabular}{|m{0.21\textwidth}|m{0.21\textwidth}|} % Adjust column width as needed
% \hline
% State 0 & State 1 \\
% \hline
% \includegraphics[width=0.1\textwidth]{figures/state_table/s0_artificial.jpeg} 
% \includegraphics[width=0.1\textwidth]{figures/state_table/s0_real.jpeg}
% & 
% \includegraphics[width=0.1\textwidth]{figures/state_table/s1_artificial.jpeg} 
% \includegraphics[width=0.1\textwidth]{figures/state_table/s1_real.jpeg}\\
 
% \hline
% State 2 & State 3 \\
% \includegraphics[width=0.1\textwidth]{figures/state_table/s2_artificial.jpeg} 
% \includegraphics[width=0.1\textwidth]{figures/state_table/s2_real.jpeg}
% & 
% \includegraphics[width=0.1\textwidth]{figures/state_table/s3_artificial.jpeg} 
% \includegraphics[width=0.1\textwidth]{figures/state_table/s3_real.jpeg}\\

% \hline
% State 4 & State 5 \\
% \includegraphics[width=0.1\textwidth]{figures/state_table/s4_artificial.jpeg} 
% \includegraphics[width=0.1\textwidth]{figures/state_table/s4_real.jpeg}
% & 
% \includegraphics[width=0.1\textwidth]{figures/state_table/s5_artificial.jpeg} 
% \includegraphics[width=0.1\textwidth]{figures/state_table/s5_real.jpeg}\\

% \hline
% \end{tabular}
% \caption{States with Images}
% \label{tab:states}
% \end{table}

State 0 served as a baseline scenario, presenting a single, unobstructed strawberry, a trivial case for direct pick algorithms. Intentionally, no human demonstration data for State 0 was included in training, making it a generalization sanity check to assess policy performance on a novel, easy task. Surprisingly, even in this simplified setting, performance was not universally high; even our best policy achieved only a 70\% success rate. This outcome underscores the challenges that relying solely on imitation learning may not guarantee robust generalization even to seemingly trivial variations of the task.

\begin{figure}
\centering
    \includegraphics[width=0.48\textwidth]{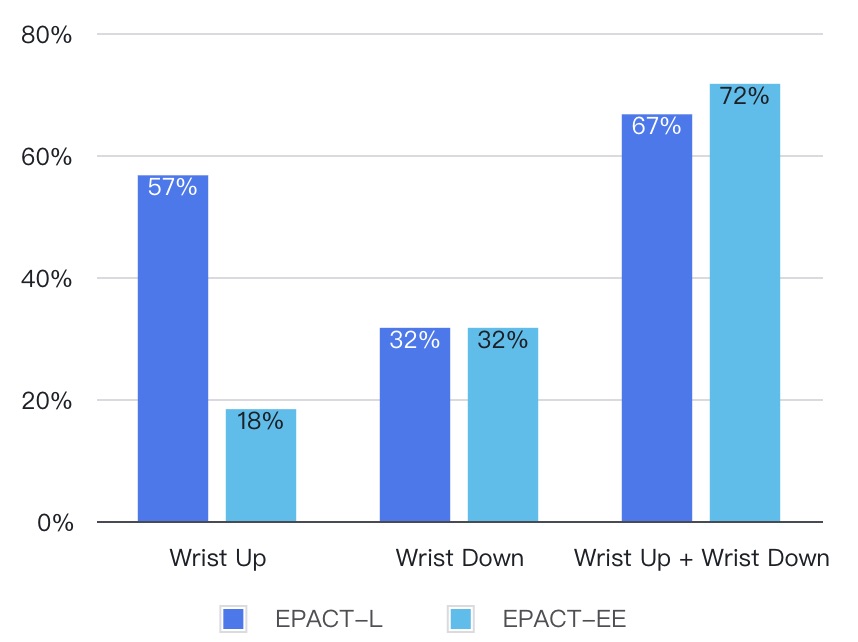}
    \caption{Comparison of performance of different camera views}
\label{fig:successful_rate_cams}
\end{figure}

\begin{figure*}
\centering
    \includegraphics[width=0.6\textwidth]{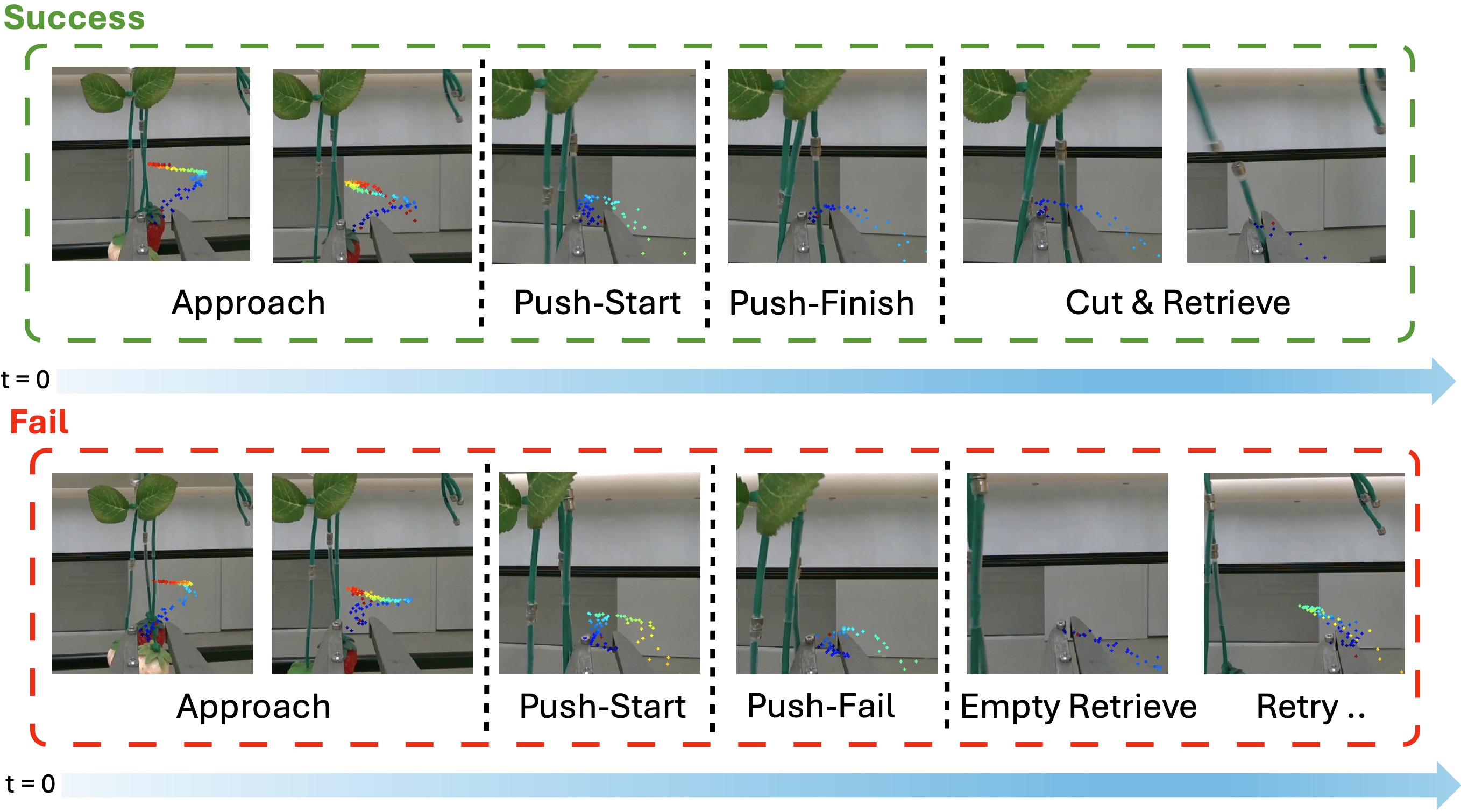}
    \caption{Top: trajectory of a successful clustered strawberry picking; Down: trajectory of a failed clustered strawberry picking; The color trajectory in each view is the current prediction of the future action visualized in image space}
\label{fig:success_and_fail}
\end{figure*}

For the clustered states, which demand dexterity for successful picking,  Figure \ref{fig:success_and_fail} (Top) illustrates representative robot trajectories during successful task executions. Both the observed success rates and these trajectory visualizations demonstrate the substantial potential of imitation learning policies to acquire human-like dexterity. This enables the robot to effectively maneuver around or even push through obstacles to successfully pick occluded strawberries. Such tasks are exceptionally difficult to program conventionally, requiring millimeter-scale precision in 3D reconstruction, complex trajectory planning within cluttered environments, and robust interaction with deformable plant materials. 

While promising, imitation learning policies still exhibited failures in clustered strawberry picking, as detailed in Figure \ref{fig:success_and_fail} (Down). These failure cases can be broadly categorized as: 1) Target Misidentification: erroneously picking a non-target strawberry; 2) Multi-Picking: inadvertently grasping and picking multiple strawberries simultaneously; and 3) Trajectory Errors: executing inaccurate approach paths that result in a failed pick attempt.

\begin{table}[htbp]
    \centering
    \captionsetup{justification=centering} 
    \caption{Comparison of different methods across states}
    \label{tab:state_comparison}
    \resizebox{\columnwidth}{!}{
    \begin{tabular}{lccccccc}
        \toprule
        Method & State0 & State1 & State2 & State3 & State4 & State5 & Avg. \\
        \midrule
        ACT & 10.0\% & 60.0\% & 30.0\% & 70.0\% & 30.0\% & 10.0\% & 35.0\% \\
        EPACT-L*& 50.0\% & \textbf{90.0}\% & 70.0\% & 50.0\% & \textbf{80.0}\% & 60.0\% & 66.7\% \\
        EPACT-EE*& \textbf{70.0}\% & 60.0\% & \textbf{80.0}\% & \textbf{90.0}\% & 60.0\% & \textbf{70.0}\% & \textbf{71.7}\%\\
        \bottomrule
    \end{tabular}
    }  % Close the resizebox after the table
    \begin{tablenotes}
        \item[*] *methods proposed by this paper.
    \end{tablenotes}
\end{table}

\subsubsection{Impact of Camera Views on Performance}
To assess the influence of camera input on visuomotor picking performance, we trained and deployed EPACT variants using different camera views.  Specifically, for each EPACT variant, we trained separate policies using two distinct monocular camera setups.  These monocular policies were then compared to policies trained with full camera observation, all of which, except for camera configuration, were trained using the same dataset and training parameters as the full-camera policies.

As illustrated in Figure \ref{fig:successful_rate_cams}, the results clearly demonstrate superior policy performance when using full camera views as input.  This improvement is likely due to the richer visual information provided by the multi-view setup.  Specifically, access to both upside and downside views (Figure \ref{fig:wrist_cameras}) enables the controller to better assess occluded conditions.  Furthermore, the multi-view perspective may afford the policies a form of stereo perception, enhancing their understanding of 3D geometry and consequently, strawberry-picking accuracy.  However, this performance gain comes with increased computational demands.  Table \ref{tab:policy_comparison} indicates that full-camera policies exhibit higher parameter counts, GPU memory usage, and inference times.  These computational costs are significant considerations for field robots operating under resource constraints and demanding real-time performance.

The study also revealed that the optimal monocular camera view is architecture-dependent. Specifically, EPACT-L achieved higher success rates with the wrist-up camera compared to the wrist-down camera. In contrast, EPACT-EE exhibited the opposite trend, performing less effectively with the wrist-up view. This divergence is likely due to differences in how EPACT-L and EPACT-EE process visual information. Each architecture may prioritize different aspects of the visual scene, and consequently, benefit from distinct camera perspectives for handling occlusions.

Despite the performance limitations relative to multi-camera setups, the successful demonstration of single monocular camera-based end-to-end strawberry picking highlights its potential for simpler and more resource-efficient robotic systems.

% \begin{table}
%     \centering
%     \caption{Comparison of different policy configurations.}
%     \label{tab:policy_comparison}
%     \resizebox{\columnwidth}{!}{
%     \begin{tabular}{lcccc}
%         \toprule
%         \shortstack{Policy\\ Name}  & Method & \shortstack{Camera\\ Used} & \shortstack{Num of\\ Param.} & \shortstack{Inference\\  Speed (ms)} \\
%         \midrule
%         EPACT-L\_W & EPACT-L & \{Wrist Up\} & 84.2M & 13.3 \\
%         EPACT-L\_WD & EPACT-L & \{Wrist Down\} & 84.2M & 13.3 \\
%         EPACT-L\_W\_WD & EPACT-L & \{Wrist Up \& Down\} & 95.3M & 23.5 \\
%         EPACT-EE\_W & EPACT-EE & \{Wrist Up\} & 84.0M & 13.1 \\
%         EPACT-EE\_WD & EPACT-EE & \{Wrist Down\} & 84.0M & 13.4 \\
%         EPACT-EE\_W\_WD & EPACT-EE & \{Wrist Up \& Down\} & 95.2M & 23.6 \\
%         \bottomrule
%     \end{tabular}
%     }
% \end{table}

\begin{table}
    \centering
    \caption{Comparison of different policy configurations.}
    \label{tab:policy_comparison}
    \resizebox{\columnwidth}{!}{
    \begin{tabular}{lcccc}
        
        \midrule
        \shortstack{Policy\\ Name}  & \shortstack{Applied\\ Method} & \shortstack{Wrist Camera\\ Used} & \shortstack{Num of\\ Param.} & \shortstack{Inference\\  Speed (ms)} \\
        \midrule
        EPACT-L\_W & EPACT-L & Up only & 84.2M & 13.3 \\
        EPACT-L\_WD & EPACT-L & Down only  & 84.2M & 13.3 \\
        EPACT-L\_W\_WD & EPACT-L & Up \& Down & 95.3M & 23.5 \\
        EPACT-EE\_W & EPACT-EE & Up only & 84.0M & 13.1 \\
        EPACT-EE\_WD & EPACT-EE & Down only & 84.0M & 13.4 \\
        EPACT-EE\_W\_WD & EPACT-EE & Up \& Down & 95.2M & 23.6 \\
        \bottomrule
    \end{tabular}
    }
\end{table}

\subsubsection{Real Clustered Strawberry Picking Transfer}
\begin{figure}
\centering
    \includegraphics[width=0.49\textwidth]{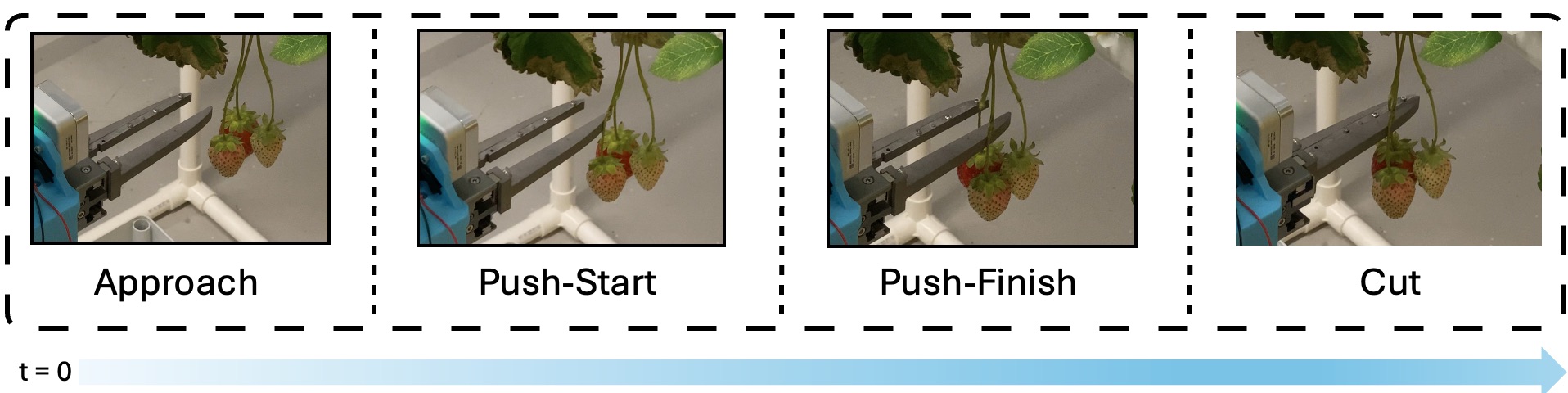}
    \caption{Trajectory of a successful real clustered strawberry picking}
\label{fig:real_cluster_picking}
\end{figure}

While our primary evaluation focused on controlled artificial strawberry scenarios, we conducted preliminary investigations into transferring the learned policies to real clustered strawberry picking. We observed the current success rate remains limited,  possibly due to challenges in bridging the simulation-to-reality gap but we do observe several times of successful picking. Figure \ref{fig:real_cluster_picking}  demonstrates a successful execution where the EPACT-EE policy successfully navigates through stems to pick a clustered strawberry. This preliminary result suggests that while significant challenges remain, visuomotor policies learned through demonstration can indeed acquire fundamental skills transferable to real agricultural environments.

\section{Discussion}

In this work, we present the use of a visuomotor policy for clustered strawberry picking. We formulate the strawberry picking problem as a future action prediction problem based on current camera and arm joint state observations. To improve end-effector pose prediction accuracy, we propose two enhanced variants of the EPACT imitation policy architecture.  We validated our approach using a 4-DOF SCARA robot equipped with a tele-operation interface for data acquisition, and rigorously evaluated performance in a controlled setting featuring six challenging strawberry occlusion scenarios.

Experimental results show that the EPACT-EE method can achieve a 72\% success rate in picking various forms of clustered strawberries, showing human-level dexterity in pushing or detouring around soft obstacles to reach the target strawberry, which is very hard to achieve using the traditional perceive-plan-pick pipeline. %with the challenges in accurate millimeter-scale 3D reconstruction, trajectory planning in clustered environments, and robot interaction with deformable plants.

% However, the application of imitation learning for clustered strawberry picking exists limitations. Generalization remains a significant challenge since environmental and light condition changes, along with countless real cluster forms, can potentially lead to picking failures. Despite collecting 511 demonstrations to train the policy, a substantial amount of additional demonstration data is still required to train a policy with a broader generalization, particularly for real field picking. One possible solution is to increase data collection efficiency by designing a portable and distributable human-picking data collection device along with its corresponding imitation learning algorithm. Another approach is to leverage simulation to obtain extensive data for pretraining the policy before fine-tuning it with field data. Pretrained robot foundation models are also a promising avenue that can potentially enhance learning to pick. While traditional perceive-plan-pick methods struggle with clustered strawberry picking, they remain more robust and efficient when picking standalone strawberries. A hybrid system could be developed to allocate the robust and fast traditional method alongside the imitation learning-based dexterous picking method, based on the condition of the target strawberry.
Despite its promise, imitation learning for clustered strawberry picking has notable limitations. Generalization remains a significant challenge because variations in environmental conditions, lighting, and the myriad forms of real strawberry clusters can lead to picking failures. Although we collected 511 demonstrations to train our policy, much more data is needed to achieve robust generalization, especially for in-field applications. To address these challenges, one potential solution is to increase data collection efficiency by developing a portable, distributable human-picking data collection device paired with the imitation learning algorithm. Alternatively, extensive simulated data can be used to pre-train the policy, followed by fine-tuning with real field data. Additionally, pre-trained robot foundation models offer a promising avenue for enhancing the learning-to-pick capability. It is also worth noting that while traditional perceive-plan-pick methods struggle with clustered strawberry picking, they remain more robust and efficient for non-clustered strawberries. A hybrid system that dynamically allocates the traditional method for simple cases and the imitation learning-based approach for dexterous, clustered scenarios could combine the strengths of both strategies.

\bibliographystyle{IEEEtran}
\bibliography{IEEEabrv,reference}  

\end{document}